\pdfoutput=1

\documentclass[11pt]{article}

\usepackage[]{ACL2023}

\usepackage{times}
\usepackage{latexsym}

\usepackage[T1]{fontenc}

\usepackage[utf8]{inputenc}

\usepackage{microtype}

\usepackage{inconsolata}
\usepackage{caption}
\usepackage{subcaption}
\usepackage{etoolbox}
\usepackage{appendix}
\usepackage{dirtytalk}
\usepackage{graphicx}
\usepackage{multirow}
\usepackage{amsmath,bm}
\usepackage{booktabs}

\title{Controlling the Extraction of Memorized Data\\from Large Language Models via Prompt-Tuning}

\makeatletter
\renewcommand*{\@fnsymbol}[1]{\ensuremath{\ifcase#1\or *\or \dagger\or \ddagger\or
   \mathsection\or \mathparagraph\or \|\or **\or \dagger\dagger
   \or \ddagger\ddagger \else\@ctrerr\fi}}
\makeatother

\author{\\
{\bf Mustafa Safa Ozdayi$^{1}$\Thanks{\enspace Work done while the author was an intern at Amazon; \texttt{mustafa.ozdayi@utdallas.edu}}, Charith Peris$^{2}$\Thanks{\enspace  \texttt{perisc@amazon.com}}, Jack Fitzgerald$^{2}$, Christophe Dupuy$^{2}$,} \\
{\bf Jimit Majmudar$^{2}$, Haidar Khan$^{2}$, Rahil Parikh$^{2}$, Rahul Gupta$^{2}$} \\
$^{1}$Department of Computer Science, The University of Texas at Dallas \\
$^{2}$Alexa AI, Amazon
}

\begin{document}
\maketitle
\begin{abstract}
Large Language Models (LLMs) are known to memorize significant portions of their training data. Parts of this memorized content have been shown to be extractable by simply querying the model, which poses a privacy risk. We present a novel approach which uses prompt-tuning to control the extraction rates of memorized content in LLMs. We present two prompt training strategies to increase and decrease extraction rates, which correspond to an attack and a defense, respectively. We demonstrate the effectiveness of our techniques by using models from the GPT-Neo family on a public benchmark. For the 1.3B parameter GPT-Neo model, our attack yields a $\bm{9.3}$ percentage point increase in extraction rate compared to our baseline. Our defense can be tuned to achieve different privacy-utility trade-offs by a user-specified hyperparameter. We achieve an extraction rate reduction of up to $\bm{97.7\%}$ relative to our baseline, with a perplexity increase of $\bm{16.9\%}$.
\end{abstract}

\section{Introduction} \label{sec:intro}

Pretrained large language models (LLMs; \citealp{devlin-etal-2019-bert}; \citealp{Radford2019LanguageMA}; \citealp{T5}; \citealp{Soltan2022}), commonly trained on massive crowd-sourced corpora, have been of much interest in the recent past due to their usage as backbones in state-of-the-art models across multiple downstream NLU tasks.
However, they have been shown to memorize significant portions of their training data that can be extracted using appropriately-crafted prompts \citep{Carlini2020ExtractingTD, Carlini2022QuantifyingMA, Zhang2021CounterfactualMI}. Such extractions pose a privacy risk to the contributors of the training data. 

In this context, methods that allow developers to control the extractability of memorized examples from LLMs are of much value. For example, methods that increase extraction rates correspond to attacks in an adversarial setting, and provide developers with the ability to analyze privacy-risk. Methods that decrease extraction rates, referred to as defenses, are useful for protecting against such attacks. Historically, defense methods tend to be compute intensive \citep{Abadi2016DeepLW, Dupuy2021AnED}.


In this work, we train continuous \emph{soft-prompts} (\citealt{lester-etal-2021-power}; hereafter referred to simply as \emph{prompts}) and leverage them as a way of passing an external signal into an LLM, to control the extraction of memorized data. We freeze the model weights, and only use the trained prompt to control the generation. First, we train prompts in an attack setting and study the extent of extractable memorized content in our models. Second, we explore a defense setting where we create prompts that reduce extraction rates and achieve different privacy-utility trade-offs, via a user-specified hyperparameter. Since the original model weights are frozen in both these settings, our methods are compute efficient across the board.

To the best of our knowledge, our work is the first to adapt the use of instructive prompts for the analysis and mitigation of privacy in LLMs. We have released the code developed for our experiments\footnote{\hyperlink{https://github.com/amazon-science/controlling-llm-memorization}{https://github.com/amazon-science/controlling-llm-memorization}}.

\section{Background and Related Work}\label{sec:background}
Previous work has shown that LLMs display memorization and has explored a range of methods that quantify extractability \citep{Carlini2018TheSS, Carlini2020ExtractingTD,Carlini2022QuantifyingMA}.  Differentially-private training~\citep{Dwork2006DifferentialP, Abadi2016DeepLW} is a popular method that has been used to mitigate this risk. However, it tends to reduce model utility and requires re-training of the LLM, which might not be feasible due to heavy computational burden. 


The use of instructive prompts for language models has been extensively researched, including use during pretraining~\citep{T5}, as a second stage of training \citep{sanh2022multitask, https://doi.org/10.48550/arxiv.2109.01652}, and during inference to guide model output \cite{NEURIPS2020_1457c0d6}. Within the third category, in order to improve upon manual prompt engineering researchers have implemented methods to learn discrete natural language prompts \citep{autoprompt:emnlp20}, to mine them \citep{jiang-etal-2020-know}, or, neglecting natural language, to learn continuous prompts \cite{li-liang-2021-prefix, lester-etal-2021-power}.

Our work leverages continuous prompts as a way of passing an external signal to a model to trigger a desired model behavior (i.e., less or more memorized data in open language generation, which map to an extraction attack and defense, respectively).
\section{Method}\label{sec:method}
\begin{figure}
\includegraphics[width=1.0\columnwidth]{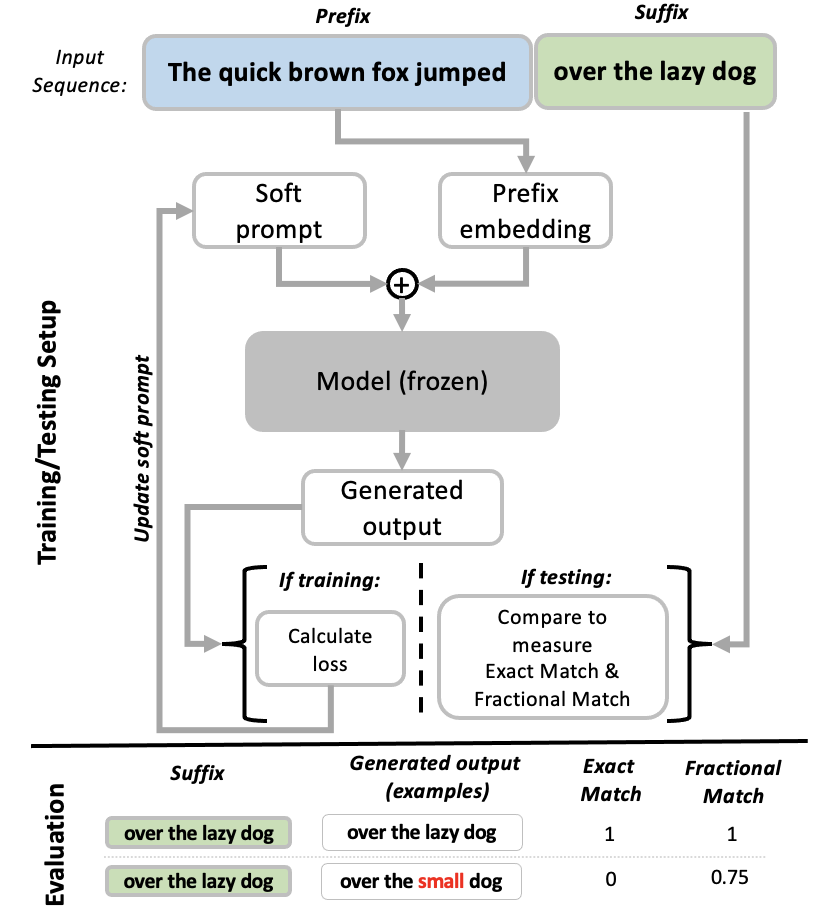}
\caption{A schematic of our setup. The upper section shows our training and testing setup while the lower section shows our evaluation metrics.} 
\label{fig:setupFig}
\end{figure}

Prompt-tuning requires the prepending of a prompt to the prefix embedding and access to the training loss (see Figure~\ref{fig:setupFig}). Given these constraints, we explore a white-box attack where the adversary has access to the target model parameters, and a black-box defense where the adversary interacts with the target model via an API. We therefore do not test our defense against our own attack.

Let [prefix || suffix] be a sequence in the training set where the prefix is of length $k$ tokens.~\citet{Carlini2022QuantifyingMA} defined a suffix to be \emph{k-extractable} if the model generates the suffix exactly, after being prompted with its the corresponding length-$k$ prefix. Our white-box attack aims to increase the number of $k$-extractable sequences, while our black-box defense aims to reduce the number of $k$-extractable sequences that can be extracted by an adversary who submits prefixes via an API.


\subsection{Attack}
\label{sec:attack}
In the attack setting, we assume that the adversary has a set of [ prefix || suffix ] sequences $S_{train}$, sampled from the training set of the target model. Their goal is to extract the suffixes corresponding to a disjoint set of prefixes, denoted by $S_{test}$\footnote{For simplicity, we assume all prefixes are $k$-length. This can easily be ensured by padding or truncating different length prefixes if needed in a real-world setting.}.

To do so, the adversary first initializes a prompt: a continuous set of $l \times e$ parameters where $e$ is the embedding size of the model, and $l$ is the length of the prompt, a hyperparameter decided by the adversary.
The prompt is trained over $S_{train}$ to facilitate the correct generation of suffixes. To do this, we first prepend the prompt to the embedding of the prefix and pass the joint embedding through the model for generation. We then minimize the loss objective (see below) with respect to the prompt while keeping the parameters of the model frozen. 

We explore two loss objectives. The first is causal language modeling (hereafter referred to as \emph{CLM}), where we minimize the cross-entropy loss over the entire sequence \citep{Radford2019LanguageMA}. In the second, the prompt is optimized by minimizing the cross entropy loss of only the suffixes, given the prefixes. Here, the training is aligned with our inference task such that during training the model is penalized only on the suffix tokens; hence we refer to it as \emph{aligned CLM}. During inference, the learned prompt is prepended to each embedding of the prefixes in $S_{test}$, and the joint embedding is passed to the model for generation (see Figure~\ref{fig:setupFig}).

\subsection{Defense}
\label{sec:defense}
In the defense setting, the defender (API owner) trains the prompt, and prepends it to the incoming prefixes before passing them to the model. Our algorithm is inspired by machine-unlearning literature~\citep{kadhe_unlearning}, and defenses against membership inference and backdoor attacks~\citep{Chen2022RelaxLossDM, Ozdayi_Kantarcioglu_Gel_2021}. We introduce a hyperparameter named \emph{learning threshold} denoted by $\theta$. During prompt training (see Section~\ref{sec:attack}), when loss is {\it less} than $\theta$ we do \emph{gradient ascent} to penalize the prompt. If the loss is {\it greater} than $\theta$, we perform gradient descent with respect to the prompt as usual. Training is stopped once the average epoch loss is equal or above $\theta$. This allows us to increase training loss in a controlled manner and stabilize it around $\theta$. Through this process, we can achieve various privacy-utility trade-offs efficiently without re-training any part of the model. To explore $\theta$, we set the initial value to be slightly above the model training loss and increase in steps of $0.25$ until desired performance is achieved.
\begin{figure*}
\includegraphics[width=1.0\textwidth]{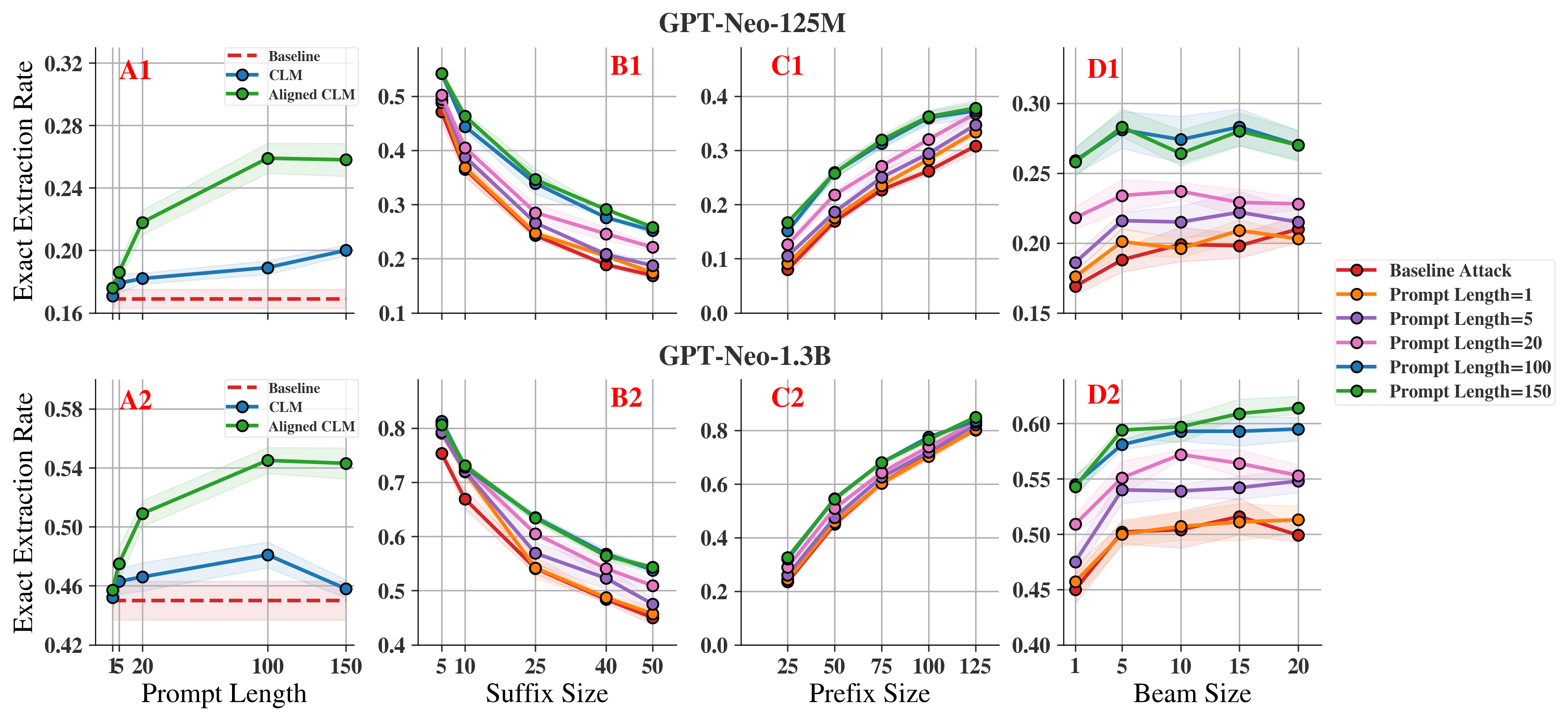}
\caption{The change in exact extraction rates against prompt length (\ref{fig:attackFig}-A1, \ref{fig:attackFig}-A2), suffix size (\ref{fig:attackFig}-B1, \ref{fig:attackFig}-B2), prefix size (\ref{fig:attackFig}-C1, \ref{fig:attackFig}-C2) and beam size (\ref{fig:attackFig}-D1, \ref{fig:attackFig}-D2). Top panels show the GPT-Neo-125M results while the bottom panels show GPT-Neo-1.3B results. The transparent polygons about each line represent 95\% confidence intervals across the points.} 
\label{fig:attackFig}
\end{figure*}

\section{Experiments}
\label{sec:exps}
For our experiments, we use the 125M and 1.3B parameter variants of the GPT-Neo models~\citep{gpt-neo}. These are public, decoder-only transformer models~\citep{Vaswani2017AttentionIA} trained using CLM on the Pile dataset~\citep{Gao2020ThePA}. We extract $S_{train}$ and $S_{test}$ from the Language Model Extraction Benchmark dataset~\citep{google-research}. This dataset contains 15k sequences sampled from the training split of the Pile where each sequence is partitioned into a prefix and suffix. In the default evaluation setting, both prefix and suffix consist of 50 tokens. We ensure a random train/test split of 14k/1k samples. 

Our evaluation metric of choice is \emph{Exact extraction rate} which is the fraction of correctly generated suffixes (i.e., all tokens of the generated suffix match with ground-truth suffix) over the test set. We additionally discuss fractional extraction rate and present results in Appendix~\ref{sec:fracExtract}. As a baseline, we use the attack analyzed in~\citet{Carlini2022QuantifyingMA}, which consists of feeding the prefixes to the model, and generating suffixes with greedy decoding. This is the only extraction attack for this setting apart from our work, to the best of our knowledge. Our training setup is discussed in Appendix~\ref{app:setup}. All experiments are repeated over 5 runs with a new random train/test split in each run.

\subsection{Attack}
We explore the performance of our attack across several dimensions: prompt length, suffix size, prefix size, and beam size. We use greedy-decoding in all cases, except the beam size experiments.

\paragraph{Prompt Length} First, we explore prompt length in the context of the default setting (prefix and suffix consist of 50 tokens; Figures~\ref{fig:attackFig}-A1 and~\ref{fig:attackFig}-A2). We note that prompts tuned with both CLM and aligned CLM provide improvements over the baseline in all cases, with aligned CLM providing the best performance. \emph{Given this, we train prompts using the aligned CLM objective for all other experiments, including our defense}. 

With aligned CLM, we achieve the highest extraction rates of $\bm{25.8\%}$ and $\bm{54.3\%}$ for the 125M and 1.3B models, respectively (an improvement of $\bm{8.9}$ and $\bm{9.3}$ percentage points, respectively), with a 100 token prompt (blue line). We observe that extraction rates increase with prompt length and tend to saturate after prompt length 100. Over-fitting was ruled out as a potential cause of saturation as there is no increase in test loss observed during training. This suggests that there is a max limit on the parameter count in the prompt that might add value for extraction purposes given our objective. We note that more sophisticated training strategies (designing better loss functions, better prompt initialization etc.) might yield better extraction rates.



\paragraph{Suffix Size}
Next, we fix the prefix size to 50 and vary the suffix size. As shown in Figures~\ref{fig:attackFig}-B1 and~\ref{fig:attackFig}-B2, extraction rates decrease roughly exponentially with suffix size. We note that as suffix size increases, longer prompts ($\ge 20$) provide greater improvements over the baseline. For example, with a prompt length of 100 (blue line) using the 1.3B model, at suffix size 5 we observe an extraction rate increase of $\bm{5.3}$ percentage points. Whereas at suffix size 50, the increase is $\bm{9.3}$ percentage points. 



\paragraph{Prefix Size}
Next, we fix the suffix size to 50 and vary the prefix size. As shown in Figures~\ref{fig:attackFig}-C1 and~\ref{fig:attackFig}-C2, extraction rates increase roughly logarithmically (as in~\citealt{Carlini2022QuantifyingMA}). Contrary to suffix size, we observe that the gaps between baseline and attacks decrease with increasing prefix size. This suggests that our attack stands to benefit a less informed adversary (small prefix sizes) when compared to the baseline.

\paragraph{Beam Decoding}
Finally, we utilize the default setting with prefix and suffix sizes at 50 tokens and vary the beam size (beam size=1 corresponds to greedy decoding). The results are shown in Figures~\ref{fig:attackFig}-D1 and~\ref{fig:attackFig}-D2. We observe that extraction rates increase across the board when increasing beam size from 1 to 5. 
However, improvements tend to plateau or oscillate when beam size is greater than 5. The 1.3B model benefits more from increasing beam size achieving the highest extraction rate of $\bm{61.4\%}$, at a beam size of 20 (with a prompt length of 150). The highest extraction rate achieved for the 125M model was $\bm{28.3\%}$ at a beam size of 15 (with a prompt length of 100).

\begin{table}[]
\setlength{\tabcolsep}{7.5pt}
\small
\begin{tabular}{llll}
\toprule
\multicolumn{1}{c}{\textbf{Model}}                                               & \multicolumn{1}{c}{\textbf{$\theta$}} & \multicolumn{1}{c}{\textbf{\begin{tabular}[c]{@{}c@{}}Exact Extract \\ Rate\end{tabular}}} & \multicolumn{1}{c}{\textbf{\begin{tabular}[c]{@{}c@{}}Pile Test \\ PPL\end{tabular}}} \\ \midrule
\multirow{4}{*}{\textbf{\begin{tabular}[c]{@{}l@{}}GPT-Neo\\ 125M\end{tabular}}} & 0$^{*}$                        & 0.169 ± 0.007                                                                              & 15.71 ± 0.431                                                                         \\
                                                                                 & 1.25                           & 0.031 ± 0.005                                                                              & 16.601 ± 0.197                                                                        \\
                                                                                 & 1.5                            & 0.006 ± 0.001                                                                              & 17.499 ± 0.156                                                                        \\
                                                                                 & \textbf{1.75}                           & \textbf{0.001 ± 0.0}                                                                                & \textbf{19.691 ± 0.598}                                                                        \\ \midrule
\textbf{\begin{tabular}[c]{@{}l@{}}GPT2\\ 124M\end{tabular}}                     & -                              & 0.004 ± 0.002                                                                              & 30.323 ± 1.019                                                                        \\ \toprule
\multirow{4}{*}{\textbf{\begin{tabular}[c]{@{}l@{}}GPT-Neo\\ 1.3B\end{tabular}}} & 0$^{*}$                        & 0.450 ± 0.015                                                                              & 9.213 ± 0.232                                                                         \\
                                                                                 & 0.5                            & 0.108 ± 0.02                                                                               & 9.758 ± 0.245                                                                         \\
                                                                                 & 0.75                           & 0.022 ± 0.004                                                                              & 10.267 ± 0.094                                                                        \\
                                                                                 & \textbf{1}                              & \textbf{0.01 ± 0.002}                                                                               & \textbf{10.775 ± 0.248}                                                                        \\ \midrule
\textbf{\begin{tabular}[c]{@{}l@{}}GPT2\\ 1.5B\end{tabular}}                     & -                              & 0.019 ± 0.002                                                                              & 17.155 ± 0.545                                                                        \\ \bottomrule
\end{tabular}
\caption{Exact extraction rates and corresponding perplexities for our defense setting, with different values of $\theta$. Values are reported as mean ± std. Extraction rates that are smaller than the corresponding GPT2 varient of similar size, achieved while perplexity values are also smaller, are good. ($^{*}\emph{no defense}$).}
\label{tab:defense}
\end{table}

\subsection{Defense}

Finally, we evaluate the privacy-utility trade-off of our black-box defense. As mentioned in Section~\ref{sec:method}, our defense is designed for a black-box adversary, and cannot be tested against our white-box attack. Therefore, we utilize the baseline attack (Section~\ref{sec:exps}) to quantify privacy. We note that longer prompts did not add value in a defense setting, so we resort to using a prompt of length 1. We utilize perplexity (PPL) on generated suffixes, to quantify the utility of the model in addition to using exact extraction rate as in Section~\ref{sec:attack}. To measure PPL, we use a random subset of 1k sequences sampled from the test split of the Pile, ensuring that PPL is measured on data unseen by the model. We also compare our metrics with those of similar sized models that were not trained on the Pile dataset (GPT2 models). Our premise here is that better performance in terms of privacy and utility, when compared to an out-of-domain model of similar size, would mean that our defense mechanism is of value to an API owner.

In Table~\ref{tab:defense}, we display our results obtained using the default evaluation setting (prefix and suffix comprise of 50 tokens). Our defense achieves lower extraction rates with competitive PPL values. For the 125M model, we achieve an exact extraction rate reduction of $\bm{99.4\%}$ relative to baseline with a PPL increase of $\bm{25.3\%}$ at $\theta=1.75$. For the 1.3B model, the extraction rate is reduced by $\bm{97.7\%}$ relative to baseline with a PPL increase of $\bm{16.9\%}$ at $\theta=1$. The ability to achieve lower extraction rates with lower PPL values as measured against the GPT2 models of the corresponding size, provides evidence that our defense is effective.
\section{Conclusion}
\label{sec:conclusion}

We present the first known effort to leverage prompt-tuning to control the extractability of memorized data from LLMs in an open language generation task. We develop a novel data extraction attack and defense, and illustrate their performance under various settings. Our attack consistently outperforms the baseline in terms of exact extraction rate. Our defense provides competitive privacy-utility trade-offs and would prove beneficial to API owners with model trained on sensitive content. These results are achieved efficiently, without any change to the original model weights. We details avenues of future work in Appendix~\ref{app:future}
\section{Limitations}\label{sec:limit}
We briefly mention some limitations of our work. First, we have only used a single dataset, and a single model family in our experiments. This is mainly due to the fact that the benchmark we use is the only publicly available dataset at this time to the best of our knowledge. We also solely focused on extraction metrics, but did not do a deeper analysis on the extracted sequences. A fine-grained analysis of extracted sequences could yield important insights for understanding memorization and extraction in LLMs. Similarly, we also did not analyze what our prompts converge to, and whether they yield explainable prompts at the time of converge. Such analysis can provide better insights as to why, for example, training prompts with aligned CLM performs better that the basic CLM setting. Finally, we believe the evaluation of our defense could be improved further by measuring other utility metrics (e.g., accuracy) on downstream tasks.

\section{Ethical Considerations}

We leverage prompt-tuning to control the extractability of memorized data from LLMs in an open language generation task and explore two settings; an attack and a defense. We acknowledge that our attack methodology could be misused by an adversary with white-box access to extract memorized private information from a target large language model. Our goal is to raise awareness in the community to the possibility and severity of this nature of attack. We hope that developers, armed with this knowledge, can use relevant defense mechanisms to avoid such potential misuse.

\section*{Acknowledgements}
The authors would like to thank Wael Hamza for helpful discussions on this topic and Stephen Rawls for help with securing the GPU instances that were required for experimentation.

\newpage
\bibliography{anthology,custom}
\bibliographystyle{acl_natbib}

\appendix

\section{Fractional Extraction Rate Results} 
\label{sec:fracExtract}

\emph{Fractional extraction rate} is the fraction of generated tokens that are both \emph{correct and in the right position}, over the dataset (see lower section of Figure~\ref{fig:attackFig}). Our reason to measure this metric is to provide a more detailed assessment of risks associated with extraction. Exact extraction rate is particularly important in cases where the attacker requires an exact match in order for the extraction to be of use; a good example is the case of extracting a credit card number. In such cases, even getting a few tokens incorrect will render the attack useless. However, when the attacker cares more about the meaning of the extracted sequences, fractional extraction rate can be a better metric to assess the risk. This is because a human might be able to infer the correct meaning of the sequence even when few tokens are wrong.

\begin{figure*}
\includegraphics[width=1.05\textwidth]{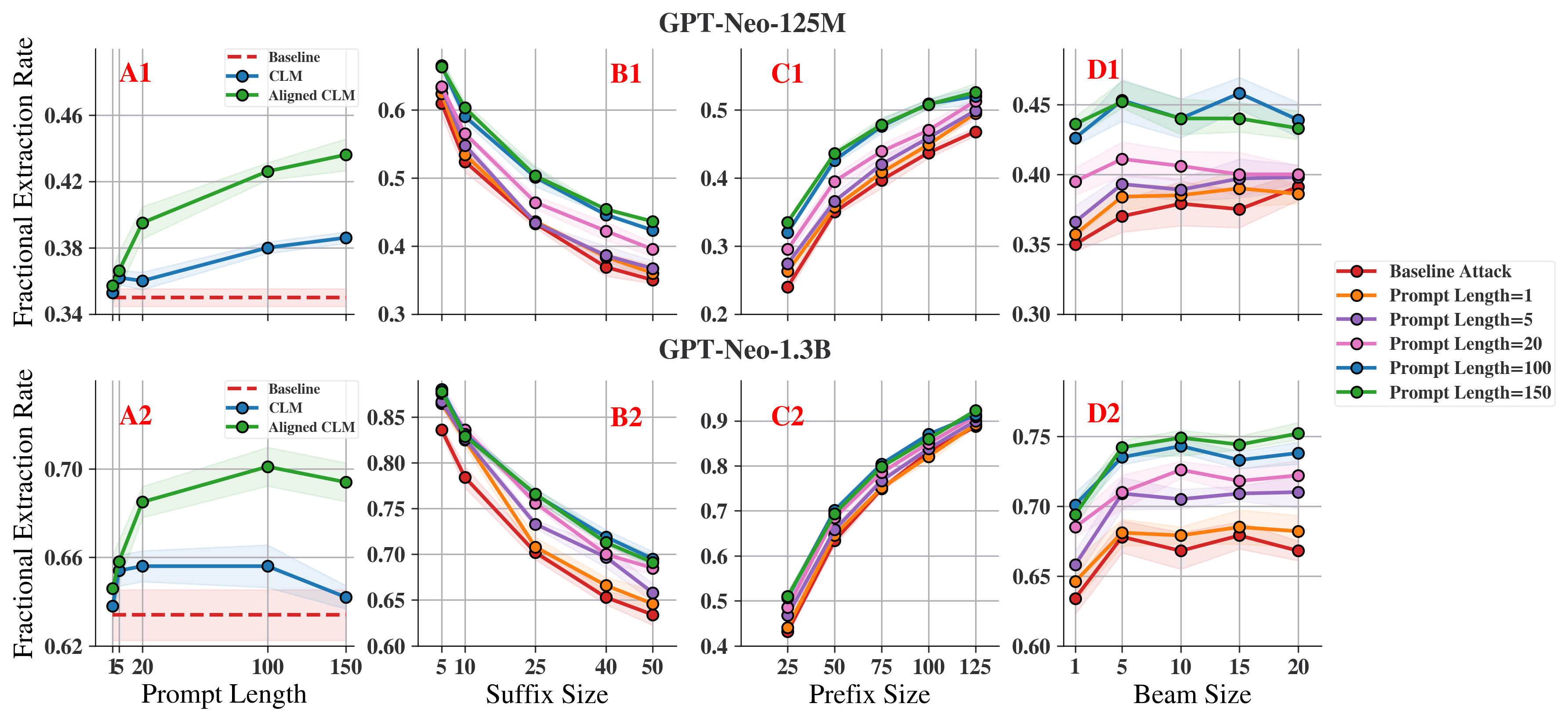}
\caption{The change in fractional extraction rates against prompt length (\ref{fig:attackFigFract}-A1, \ref{fig:attackFigFract}-A2), suffix size (\ref{fig:attackFigFract}-B1, \ref{fig:attackFigFract}-B2), prefix size (\ref{fig:attackFigFract}-C1, \ref{fig:attackFigFract}-C2) and beam size (\ref{fig:attackFigFract}-D1, \ref{fig:attackFigFract}-D2). Top panels show the GPT-Neo-125M results while the bottom panels show GPT-Neo-1.3B results. The transparent polygons about each line represent 95\% confidence intervals across the points.} 
\label{fig:attackFigFract}
\end{figure*}

The results related to this metric are shown in Figure~\ref{fig:attackFigFract}. Comparing these results with the exact extraction rate results (Figure~\ref{fig:attackFig}), we observe the same trends across all of our experiment. We note that the same shared trends are observed in the case of our defense. In this case the fractional extraction rate results are tabulated in Table~\ref{tab:defenseFract}.

\begin{table}[]
\setlength{\tabcolsep}{7.5pt}
\small
\begin{tabular}{llll}
\toprule
\multicolumn{1}{c}{\textbf{Model}}                                               & \multicolumn{1}{c}{\textbf{$\theta$}} & \multicolumn{1}{c}{\textbf{\begin{tabular}[c]{@{}c@{}}Fract Extract \\ Rate\end{tabular}}} & \multicolumn{1}{c}{\textbf{\begin{tabular}[c]{@{}c@{}}Pile Test \\ PPL\end{tabular}}} \\ \midrule
\multirow{4}{*}{\textbf{\begin{tabular}[c]{@{}l@{}}GPT-Neo\\ 125M\end{tabular}}} & 0$^{*}$                        & 0.35 ± 0.006                                                                              & 15.71 ± 0.431                                                                         \\
                                                                                 & 1.25                           & 0.192 ± 0.011                                                                              & 16.601 ± 0.197                                                                        \\
                                                                                 & 1.5                            & 0.123 ± 0.005                                                                              & 17.499 ± 0.156                                                                        \\
                                                                                 & \textbf{1.75}                           & \textbf{0.087 ± 0.003}                                                                                & \textbf{19.691 ± 0.598}                                                                        \\ \midrule
\textbf{\begin{tabular}[c]{@{}l@{}}GPT2\\ 124M\end{tabular}}                     & -                              & 0.099 ± 0.003                                                                              & 30.323 ± 1.019                                                                        \\ \toprule
\multirow{4}{*}{\textbf{\begin{tabular}[c]{@{}l@{}}GPT-Neo\\ 1.3B\end{tabular}}} & 0$^{*}$                        & 0.634 ± 0.013                                                                              & 9.213 ± 0.232                                                                         \\
                                                                                 & 0.5                            & 0.316 ± 0.022                                                                               & 9.758 ± 0.245                                                                         \\
                                                                                 & 0.75                           & 0.171 ± 0.004                                                                              & 10.267 ± 0.094                                                                        \\
                                                                                 & \textbf{1}                              & \textbf{0.128 ± 0.006}                                                                               & \textbf{10.775 ± 0.248}                                                                      \\ \midrule
\textbf{\begin{tabular}[c]{@{}l@{}}GPT2\\ 1.5B\end{tabular}}                     & -                              & 0.166 ± 0.003                                                                              & 17.155 ± 0.545                                                                        \\ \bottomrule
\end{tabular}
\caption{Fractional extraction rates and corresponding perplexities for our defense setting, with different values of $\theta$. Values are reported as mean ± std. Extraction rates that are smaller than the corresponding GPT2 varient of similar size, achieved while perplexity values are also smaller, are good.($^{*}\emph{no defense}$).}
\label{tab:defenseFract}
\end{table}

\section{Training Setup}
\label{app:setup}
Our soft-prompts are initialized to random word embeddings as described in~\citet{lester-etal-2021-power}. We use a batch size of 128 and an Adam optimizer~\citep{Kingma2014AdamAM} with a learning rate of $5e-4$. For the attack setting, the prompts are trained for 15 epochs. In the defense case, the prompts are trained until training loss stabilizes around the specified $\theta$ value (as described in Section~\ref{sec:defense}), which happens within 2-3 epochs in our experiments. 

We use a Pytorch~\citep{pytorch} implementation where we leverage the HuggingFace Accelerate~\citep{HF} and DeepSpeed~\citep{DeepSpeedcite} libraries to handle distributed training over 8 GPUs with \texttt{fp16} mixed precision. On a \texttt{p3dn.24xlarge} instance, the average attack prompt training time was $0.9$ hours per prompt while the average defense prompt training time was $0.02$ hours per prompt.

\section{Future work}
\label{app:future}
We have several avenues that we would like to explore in the context of future work. We envision that more sophisticated training strategies might yield better extraction rates in our attack setting (designing better loss objectives, better initialization of soft-prompts etc.) and we would like to explore this further.

We would like to explore different prompt learning algorithms such as other parameter-efficient training methods~\citep{li-liang-2021-prefix, Hu2021LoRALA}, and hard-prompt learning methods \citep{Wallace2019UniversalAT}, in order to conduct a more robust analysis of extraction rates.

We would like to test the transferability of trained prompts across different models and datasets. 

Finally, we would like to combine our defense with other existing defenses such as those applied at training time (e.g. versions of differentially private stochastic gradient descent;~\citealt{Abadi2016DeepLW, Dupuy2021AnED}) or those applied at decoding stage (e.g., differentially private decoding;~\citealt{Majmudar2022DifferentiallyPD}). The goal would be to achieve better privacy-utility trade-offs under a combination of such defenses.
\end{document}